%% file: main.tex
\documentclass[letterpaper, 10pt, conference]{ieeeconf} 
\IEEEoverridecommandlockouts
\usepackage[ruled,vlined,linesnumbered]{algorithm2e}
\usepackage{cite}
\usepackage{amsmath,amssymb,amsfonts,algorithmic,textcomp, multirow}
\usepackage{graphicx,subfig,float,xcolor,url,hyperref,etoolbox}
\usepackage[nolist]{acronym}
\hypersetup{
    colorlinks,
    linkcolor={blue!50!black},
    citecolor={blue!50!black},
    urlcolor={black!80!black}
}

\DeclareMathSymbol{\shortminus}{\mathbin}{AMSa}{"39}

\usepackage{tabularx}
\usepackage{soul}

\def\BibTeX{{\rm B\kern-.05em{\sc i\kern-.025em b}\kern-.08em
    T\kern-.1667em\lower.7ex\hbox{E}\kern-.125emX}}
    
\makeatletter
  \patchcmd{\@maketitle}
   {\addvspace{0.5\baselineskip}\egroup}
   {\addvspace{-0.1\baselineskip}\egroup}
   {}
   {}
\makeatother
\SetKw{KwBy}{by}
\SetKw{KwNot}{not}
\makeatletter
\newcommand\notsotiny{\@setfontsize\notsotiny\@vipt\@viipt}
\makeatother
\newcommand{\ours}{\notsotiny{SAGAT}}

\newcommand{\grasp}{g_{p_i}}
\newcommand{\graspoptimal}{g^*_p}
\newcommand{\graspregion}{G^*}

\newcommand{\task}{\mathcal{T}}
\newcommand{\policy}{\pi_\tau}
\newcommand{\effect}{\alpha_\tau}
\newcommand{\effectset}{A_\tau}
\newcommand{\effectnew}{\overline{\alpha}_\tau}

\newcommand{\actionregion}{S_O}

\newcommand{\library}{\mathcal{L}}
\newcommand{\confidencefunction}{C}

\begin{document}

\title{Self-Assessment of Grasp Affordance Transfer}
\author{Paola Ard{\'o}n$^{*}$, {\`E}ric Pairet$^{*}$, Ronald P. A. Petrick, Subramanian Ramamoorthy, and Katrin S. Lohan
\thanks{The authors are with the Edinburgh Centre for Robotics at the University of Edinburgh and Heriot-Watt University, Edinburgh, UK. This research is supported by the Scottish Informatics and Computer Science Alliance (SICSA), EPSRC ORCA Hub (EP/R026173/1) and consortium partners. \tt\{paola.ardon,eric.pairet\}@ed.ac.uk
}
\thanks{${}^{*}$These authors contributed equally to this work.}
}

\input{acronyms}
\maketitle

\begin{abstract}
Reasoning about object grasp affordances allows an autonomous agent to estimate the most suitable grasp to execute a task. While current approaches for estimating grasp affordances are effective, their prediction is driven by hypotheses on visual features rather than an indicator of a proposal's suitability for an affordance task. Consequently, these works cannot guarantee any level of performance when executing a task and, in fact, not even ensure successful task completion. In this work, we present a pipeline for \acf{SAGAT} based on prior experiences. We visually detect a grasp affordance region to extract multiple grasp affordance configuration candidates. Using these candidates, we forward simulate the outcome of executing the affordance task to analyse the relation between task outcome and grasp candidates. The relations are ranked by performance success with a heuristic confidence function and used to build a library of affordance task experiences. The library is later queried to perform one-shot transfer estimation of the best grasp configuration on new objects. Experimental evaluation shows that our method exhibits a significant performance improvement up to $11.7\%$ against current state-of-the-art methods on grasp affordance detection. Experiments on a PR2 robotic platform demonstrate our method's highly reliable deployability to deal with real-world task affordance problems.
\end{abstract}

\input{Sections/introduction.tex}
\input{Sections/related_work.tex}

\input{Sections/method.tex}
\input{Sections/results.tex}
\input{Sections/contribution.tex}
\bibliographystyle{ieeetr}
\bibliography{bibliography}
\end{document}

%% file: acronyms.tex
\begin{acronym}[ransac]
  \acro{LbD}{learning by demonstration}
  \acro{RL}{reinforcement learning}
  \acro{SVM}{Support Vector Machine}
  \acro{DOF}{degrees-of-freedom}
  \acro{CAD}{computer-aided design}
  \acro{ROI}{regions of interest}
  \acro{MRF}{Markov random fields}
  \acro{ECV}{early cognitive vision}
  \acro{IADL}{instrumental activities of daily living}
  \acro{CDR}{cognitive developmental robotics}
  \acro{2-D}{two-dimensional}
  \acro{3-D}{three-dimensional}
  \acro{RANSAC}{random sample consensus}
  \acro{RGB-D}{red-green-blue depth}
  \acro{IFR}{International Federation of Robotics}
  \acro{CNN}{convolutional neural network}
  \acro{KB}{knowledge base}
  \acro{MSE}{mean square error}
  \acro{MLN}{Markov logic network}
  \acro{XAI}{explainable artificial intelligence}
  \acro{MC-SAT}{model-constructing satisfiability calculus}
  \acro{WCSP}{weighted constraint satisfaction problem}
  \acro{MAP}{Maximum--Likelihood}
  \acro{O-CNN}{octree-based convolutional neural networks}
  \acro{OACs}{object-action complexes}
  \acro{CAD}{computer-aided-design}
  \acro{ROC}{receiver operating characteristics}
  \acro{AUC}{area under the curve}
  \acro{MCMC}{Markov chain Monte Carlo}
  \acro{FOL}{first-order logic}
  \acro{DMP}{dynamic movement primitive}
  \acro{M-RCNN}{mask RCNN}
  \acro{SAGAT}{self-assessment of grasp affordance transfer}
\end{acronym}

%% file: Sections/introduction.tex
\section{Introduction}\label{sec:intro}

Affordances have attained new relevance in robotics over the last decade~\cite{jamone2018affordances,min2016affordance}. Affordance refers to the possibility of performing different tasks with an object~\cite{Gibson77-affordances}. As an example, grasping a pair of scissors from the tip affords the task handing over, but not a cutting task. Analogously, not all the regions on a mug's handle comfortably afford to pour liquid from it. Current grasp affordance solutions successfully detect the parts of an object that afford different tasks~\cite{Lenz2015,chu2018real,chu2019learning,bohg2010learning,AffordanceNet18,ardon2019learning}. This allows agents to contextualise the grasp according to the objective task and also, to novel object instances. Nonetheless, these approaches lack an insight into the level of suitability that the grasp offers to accomplish the task. As a consequence, current literature on grasp affordance cannot guarantee any level of performance when executing the task and, in fact, not even a successful task completion.

On the grounds of the limitations mentioned above, a system should consider the expected task performance when deciding a grasp affordance. However, this is a challenging problem, given that the grasp and the task performance are codefining and conditional on each other~\cite{Montesano2008LearningOA}. Recent research in robot affordances proposes to learn this relation via trial and error of the task~\cite{fang2019learning,mandlekar2018roboturk,kroemer2012kernel}. Nevertheless, given the extensive amount of required data, the method can solely learn a single task at a time and perform on known scenarios. In contrast, an autonomous agent is expected to be capable of dealing with multiple task affordance problems even when those involve unfamiliar objects and new scenarios.

\begin{figure}[t!]
  \centering
  \includegraphics[width= 8.5cm]{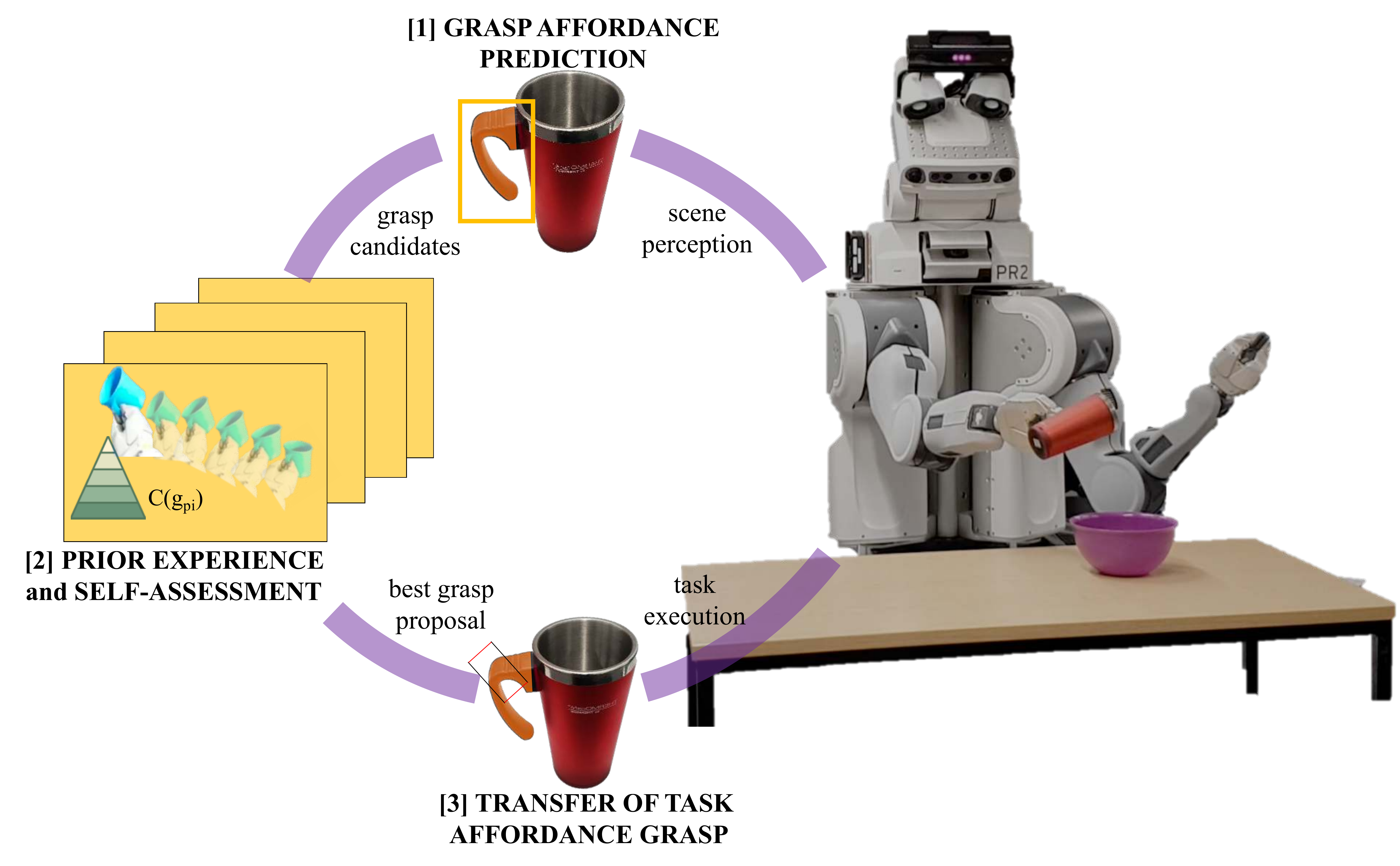}
  \vspace{-0.2cm}
  \caption{PR2 self-assessing a pouring affordance task. The system first predicts the object's grasp affordances. Then, based on prior affordance task experiences and a heuristic confidence metric, it self-assesses the new object's grasp configuration that is most likely to succeed at pouring. \label{fig:intro}}
  \vspace{-0.35cm}
\end{figure}

In this paper, we present a novel experience-based pipeline for \acf{SAGAT} that seeks to overcome the lack of deployment reliability of current state-of-the-art methods of grasp affordance detection. The proposed approach, depicted in Fig.~\ref{fig:intro}, starts by extracting multiple grasp configuration candidates from a given grasp affordance region. The outcome of executing a task from the different grasp candidates is estimated via forward simulation. These estimates are employed to evaluate and rank the relation of task performance and grasp configuration candidates
via a heuristic confidence function.
Such information is stored in a library of task affordances. The library serves as a basis for one-shot transfer to identify grasp affordance configurations similar to those previously experienced, with the insight that similar regions lead to similar deployments of the task. We evaluate the method's efficacy on addressing novel task affordance problems by training on one single object and testing on multiple new ones. We observe a significant performance improvement up to $11.7\%$ in the considered tasks when using our proposal in comparison to state-of-the-art approaches on grasp affordance detection. Experimental evaluation on a PR2 robotic platform demonstrates highly reliable deployability of the proposed method in real-world task affordance problems. 

\begin{figure*}[!hbt]
  \centering
  \includegraphics[width=17.8cm]{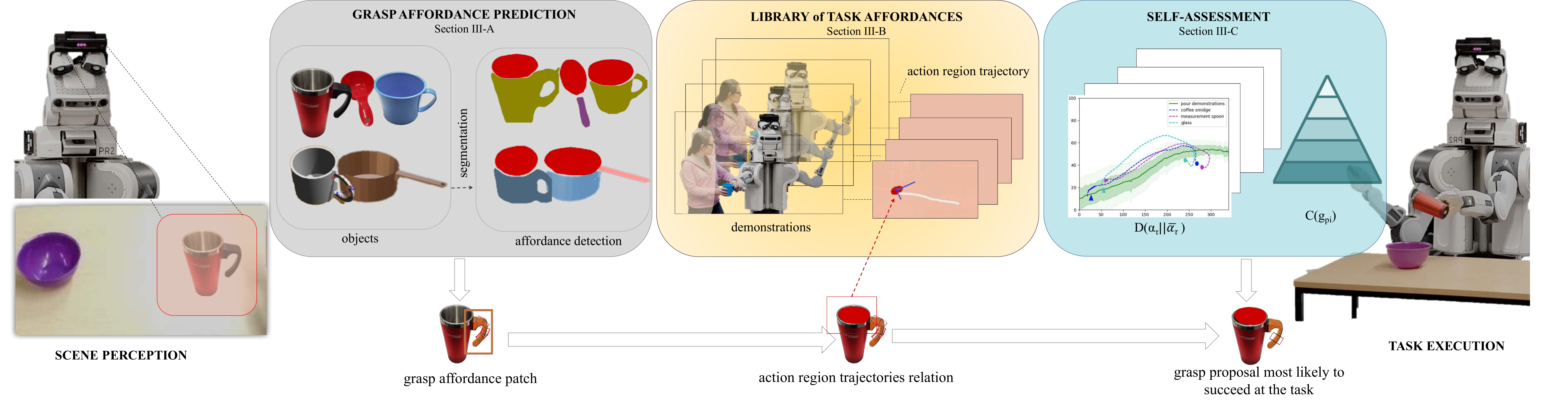}
  \caption{Proposed framework for \acl{SAGAT}. After predicting a grasp affordance region, the most suitable grasp is determined based on a library of prior task affordance experiences and a heuristic confidence metric. }\label{fig:framework}
\end{figure*}

%% file: Sections/related_work.tex
\section{Related Work}\label{sec:related_work}

Understanding grasp affordances for objects has been an active area of research for robotic manipulation tasks. Ideally, an autonomous agent should be able to identify all the tasks that an object can afford, and infer the grasp configuration that leads to a successful completion of each task. A common approach to tackle this challenge is via visual features, e.g.~\cite{Lenz2015,chu2018real,chu2019learning,bohg2010learning}. Methods based on visual grasp affordance detection identify candidate grasps either via deep learning architectures that detect grasp areas on an object \cite{Lenz2015,chu2018real,chu2019learning}, or via supervised learning techniques that obtain grasping configurations based on an object's shape \cite{bohg2010learning}. While these techniques offer robust grasp candidates, they uniquely seek grasp stability. Consequently, these methods cannot guarantee any level of performance when executing a task, and in fact, not even a successful task completion. In order to, move towards reliable task deployment on autonomous agents, there is the need to bridge the gap between grasp affordance detection and task-oriented grasping. 

\subsubsection*{\textbf{Grasp affordances}}
Work on grasp affordances aims at robust interactions between objects and the autonomous agent. However, it is typically limited to a single grasp affordance detection per object, thus reducing its deployment in real-world scenarios. Some works, such as \cite{kruger2011object}, focus on relating abstractions of sensory-motor processes with object structures (e.g., \ac{OACs}) to extract the best grasp candidate given an object affordance. Others use purely visual input to learn affordances using deep learning \cite{AffordanceNet18,chu2019learning} or supervised learning techniques to relate objects and actions \cite{song2010learning,Montesano2009LearningGA,moldovan2012learning,ardon2019learning}. Although these works are successful in detecting grasp affordance regions, they hypothesise suitable grasp configurations based on visual features, rather than indicators that hint such proposals suitability to accomplish an affordance task.

\subsubsection*{\textbf{Task affordances}}
The end goal of grasping is to manipulate an object to fulfil a goal-directed task. When the grasping problem is contextualised into tasks, solely satisfying the grasp stability constraints is no longer sufficient. Nonetheless, codefining grasp configurations with task success is still an open problem. Along this line, some works focus entirely on learning tasks where the object category does not influence the outcome, such as pushing or pulling \cite{song2010learning,moldovan2012learning}. Hence, reliable extraction of grasp configurations is neglected. Another approach is to learn grasp quality measures for task performance via trial and error \cite{fang2019learning,mandlekar2018roboturk,kroemer2012kernel}. Based on the experiences, these studies build semantic constraints to specify which object regions to hold or avoid. Nonetheless, their dependency on great amounts of prior experiences and the lack of generalisation between object instances remain to be the main hurdle of these methods.

Our work seeks to bridge the gap between grasp affordances and task performance existing in prior work. The proposed approach unifies grasp affordance reasoning and task deployment in a self-assessed system that, without the need for extensive prior experiences, is able to transfer grasp affordance configurations to novel object instances.

%% file: Sections/method.tex
\section{Proposed Method \label{sec:method}}

An autonomous agent must be able to perform a task affordance in different scenarios. Given a particular object and task $\task$ to perform, the robot must select a suitable grasp affordance configuration $\graspoptimal$ that allows executing the task's policy $\policy$ successfully. Only the correct choice of both $\graspoptimal$ and $\policy$ leads to the robot being successful at addressing the task affordance problem. Despite the strong correlation between $\graspoptimal$ and the $\policy$ execution performance, current approaches in the literature consider these elements to be independent. This results in grasping configurations that are not suitable for completing the task.

In this section, we introduce our approach to self-assess the selection of a suitable grasp affordance configuration according to an estimate of the task performance. Fig.~\ref{fig:framework} illustrates the proposed pipeline which (i)~detects from visual information a set of grasping candidates lying in the object's grasp affordance space (Section~\ref{sc:kb}), (ii)~exploits a learnt library of task affordance policies to forward simulate the outcome of executing the task from the grasping candidates (Section~\ref{sc:mrcnn_and_dmp}), and then (iii)~evaluates the grasp configuration candidates subject to a heuristic confidence metric (Section~\ref{sec:entropy}) which allows for one-shot transfer of the grasp proposal (Section~\ref{sec:one_shot_library}). Finally, in Section~\ref{sec:framework}, we detail how theses components fit in the scheme of a robotic agent dealing with task affordance problems autonomously.

\subsection{Prediction of Grasp Affordance Configurations\label{sc:kb}}

The overall goal of this work is, given an object's grasp affordance region $\graspregion$, to find a grasp configuration ${\graspoptimal}$ that allows the robot to successfully employ an object for a particular task. In the grasp affordance literature, it is common to visually detect and segment the grasp affordance region $\graspregion$ using mapping to labels~\cite{AffordanceNet18,ardon2019learning,chu2019learning}. While these methods all predict $\graspoptimal$ via visual detection hypotheses, none estimate the configuration proposals based on a task performance insight. This relational gap endangers a successful task execution. Instead, an autonomous agent should be capable of discerning the most suitable grasp that benefits the execution of a task.

\begin{figure}[t!]
        \centering
        \subfloat[$\graspregion$ from \cite{ardon2019learning} \label{fig:kb}]{\includegraphics[height=2.38cm]{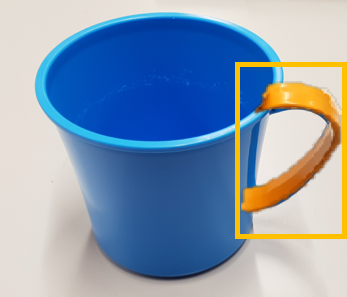}} \;\!\!
        \subfloat[$\grasp$ from \cite{chu2018real} \label{fig:deepGrasp}]{\includegraphics[height=2.38cm]{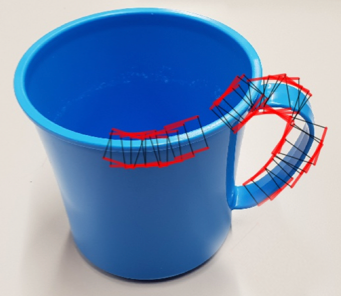}} \;\!\!
        \subfloat[Combined ${\grasp \in \graspregion}$\label{fig:both}]{\includegraphics[height=2.38cm]{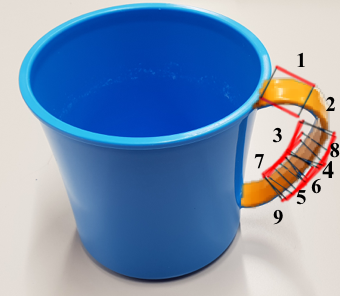}}
    
        \caption{Prediction of grasp affordance configurations for the pouring task. (a)~Patch affording the pouring task, (b)~reliable grasp configurations from DeepGrasp, (c)~pruned space for reliable grasp candidates that afford the task pouring.}
        \label{fig:examples_on_mug}
\end{figure}

To bridge this gap, in our method we consider a grasp affordance region $\graspregion$ in a generic form such as the bounding box provided by \cite{ardon2019learning} (see Fig.~\ref{fig:kb}). We are interested in pruning this region by finding multiple grasp proposal candidates. With this aim, we use the pre-trained DeepGrasp model~\cite{chu2018real}, a deep CNN that computes reliable grasp configurations on objects. The output grasp proposals $\grasp$ from DeepGrasp, which do not account for affordance relation, are shown in Fig.~\ref{fig:deepGrasp}. The pruned region (see Fig.~\ref{fig:both}), denoted as ${\grasp \in \graspregion}$, provides a set of grasp configuration candidates that accounts for both reliability and affordability.

\subsection{Library of Task Affordances \label{sc:mrcnn_and_dmp}}

The success of an affordance task $\task$ lies in executing the corresponding task policy $\policy$ from a suitable grasp configuration $\graspoptimal$. This is a difficult problem given that the $\policy$ and $\graspoptimal$ are codefining~\cite{Montesano2008LearningOA}. Namely, the task's requirements constrain the possibly suitable grasp configurations $\graspoptimal$, at the same time that the choice of $\graspoptimal$ conditions the outcome of executing the task's policy $\policy$. Additionally, determining whether the execution of a task is successful requires a performance indicator. To cope with this challenge, we build on our previous work~\cite{pairet2019learning} to 
learn a library $\library$ of task affordances from human demonstrations. The library aims at simultaneously guiding the robot on the search of a suitable task policy $\policy$ while informing about its expected outcome $\effect$ when successful. All these elements serve as the basis of the method described in Section~\ref{sec:entropy} to determine $\graspoptimal$ via self-assessment of the candidates ${\grasp \in \graspregion}$.

In this work, we build the library of task affordances as:
\begin{equation}
    \library = \bigl\{\task_1 \rightarrow \{\pi_{\tau_1}, A_{\tau_1}\}, \cdots, \task_n \rightarrow \{\pi_{\tau_n}, A_{\tau_n}\}\bigr\},
    \label{eq:library}
\end{equation}
where $\policy$ is a policy encoding the task in a generalisable form, and ${\effect \in \effectset}$ is a set of possible successful outcomes when executing $\policy$. In our implementation, $\policy$ is based on \acp{DMP}~\cite{ijspeert2013dynamical,pairet2019learningb}. \acp{DMP} are differential equations encoding behaviour towards a goal attractor. We initialise the policies via imitation learning, and use them to reproduce an observed motion while generalising to different start and goal locations, as well as task durations. 

\begin{figure}[b!]
    \centering
    \subfloat[Unsuccessful pour (grasping at $g_{p_1}\:\!\!$)]{\includegraphics[width=7.4cm]{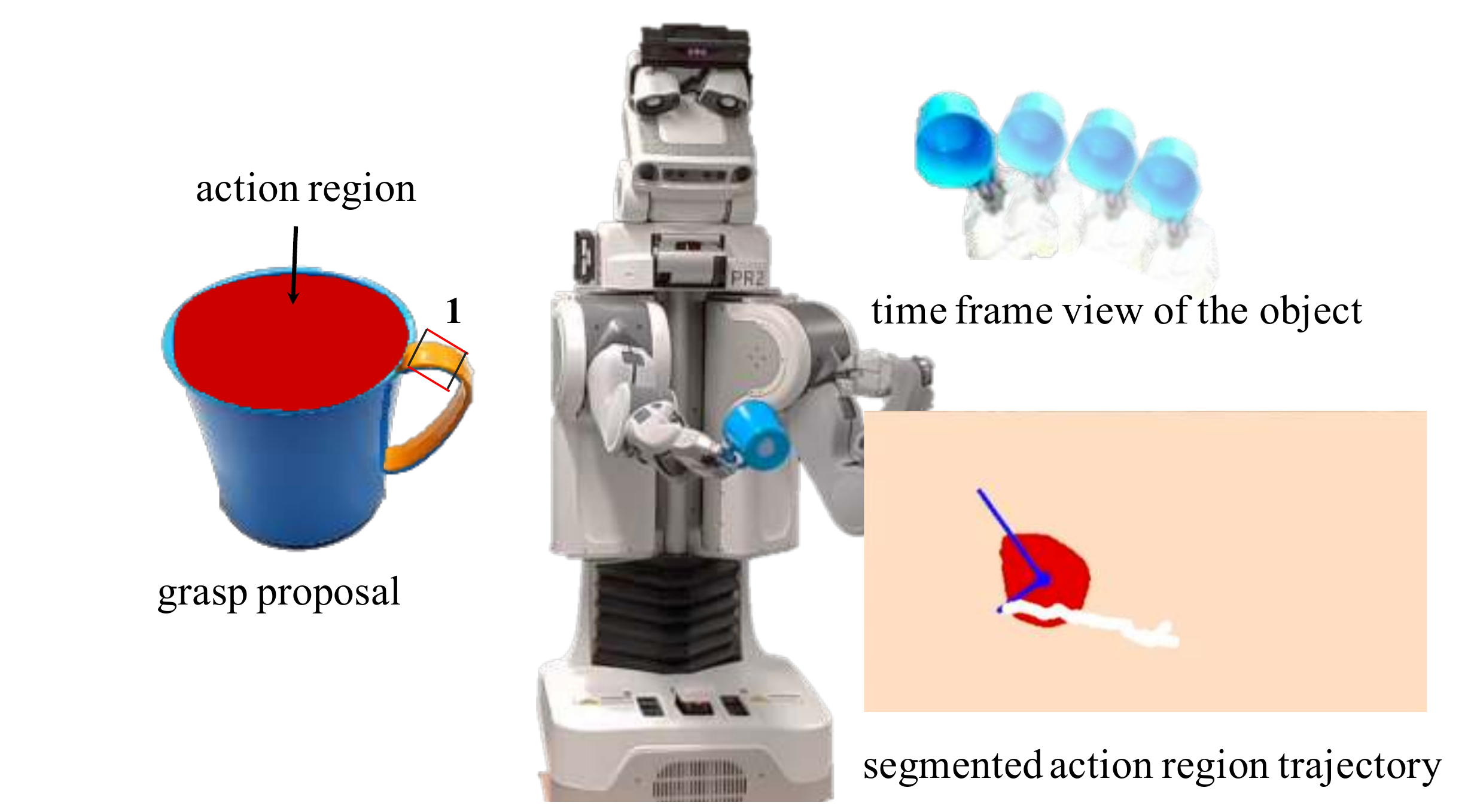}} \\
    \subfloat[Successful pour (grasping at $g_{p_2}\:\!\!$)]{\centering
    \includegraphics[width=7.4cm]{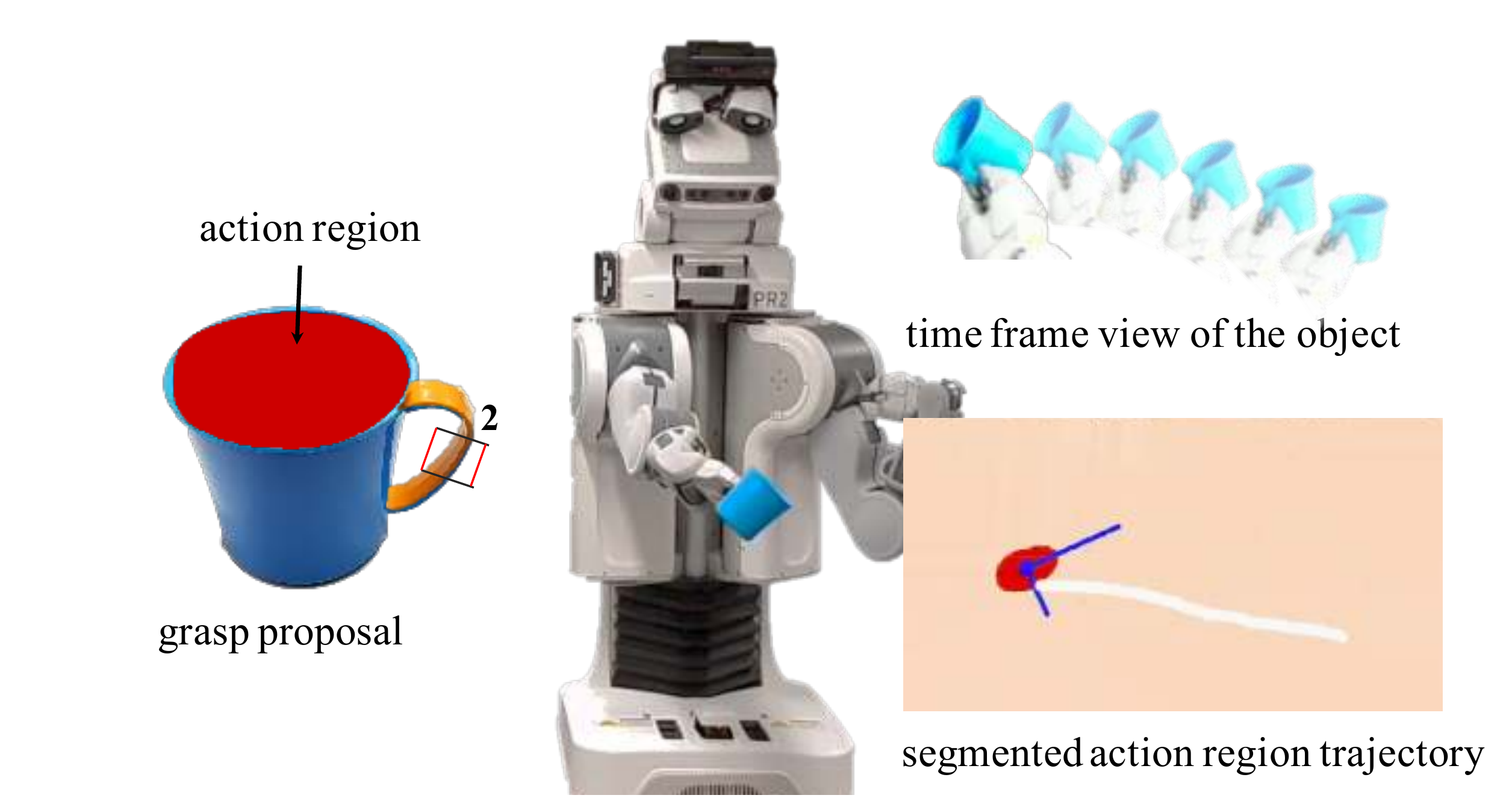}} 
    \caption{Example of a pouring task from two different grasp configurations. Each situation illustrates the raw \ac{2-D} camera input of the object and the segmented action region that affords the pouring task.
    \label{fig:grasp}}
 \end{figure}

Regarding the set of possible successful outcomes ${\effect \in \effectset}$, we provide the robot with multiple experiences. We define the outcome $\effect$ as the state evolution of the object's action region $\actionregion$ through the execution of the task. We employ \ac{M-RCNN} \cite{he2017mask} to train a model that detects objects subparts as action regions $\actionregion$. As exemplified in Fig.~\ref{fig:grasp}, the action region state provides a meaningful indicator of the task. This information is used as the basis for our confidence metric, which evaluates the level of success of an affordance task for a grasping proposal.

\subsection{Search-Based Self-Assessment of Task Affordances} \label{sec:entropy}

The task policies $\policy$ learnt in Section~\ref{sc:mrcnn_and_dmp} allow a previously experienced task from any candidate grasp ${\grasp \in \graspregion}$ to be performed. Nonetheless, executing $\policy$ from any grasp configuration may not always lead to suitable performance. For example, Fig.~\ref{fig:grasp} depicts the case where grasping the mug from $g_{p_1}$ prevents the robot from performing a pouring task as adequately as when grasping it from $g_{p_2}$.

We propose to self-assess the outcome of executing the task's policy $\policy$ from ${\grasp \in \graspregion}$ before deciding the most suitable grasp configuration $\graspoptimal$ on a new object. This is efficiently done by forward simulation of the \ac{DMP}-encoded $\policy$. From each roll-out, we look at the object's state action region $\effectnew$ as a suitable task performance indicator. To this aim, we consider the entropy between the demonstrated successful task outcomes $\effect$ and the simulated outcome $\effectnew$ in the form of Kullback-Leibler divergence \cite{perez2008kullback}:
\begin{equation}
    \label{eq:kld}
    D(\effect || \effectnew) = \sum_{i \in I} \effect(i) \log\!\left( \frac{\effect(i)}{\effectnew(i)} \right)\!,
\end{equation}
which results in a low penalisation when the forward simulated outcome $\effectnew$ is similar to a previously experienced outcome in $\effectset$, and a high penalisation otherwise. Then, we propose to rank the grasping candidates ${\grasp \in \graspregion}$ according to a confidence metric which estimates the suitability of a candidate $\grasp$ for a given $\task$ as:
\begin{equation}
    \label{eq:conf}
    \confidencefunction(\grasp) = \max_{\effect \in \effectset} D^{\shortminus1}(\effect || \effectnew).
\end{equation}

Finally, we select the grasping configuration $\graspoptimal$ among all grasping candidate ${\grasp \in \graspregion}$ as: 
\begin{equation}
    \graspoptimal = \arg \max_{\grasp \in \graspregion} \confidencefunction(\grasp) \quad s.t. \quad \confidencefunction(\grasp) > \delta, \label{eq:optimal_grasp}
\end{equation}
which returns the grasp configuration with highest confidence of successfully completing the task. This assessment is subject to a minimum user-defined confidence level $\delta$ that rejects under-performing grasp configuration proposals. As explained in the experimental setup, such a threshold is adjusted from demonstration by a binary classifier.

\subsection{One-Shot Self-Assessment of Task Affordances \label{sec:one_shot_library}}

The search-based strategy presented in Section~\ref{sec:entropy} in the grasp affordance region can be time and resource consuming if performed for every single task affordance problem. Alternatively, we propose to augment the library in~\eqref{eq:library} with an approximate of the prior experienced outcomes $\effect$ per grasp configuration $\grasp$, such that it allows for one-shot assessment. Namely, we extract the spatial transform of all experienced grasps with respect to the detected grasp affordance region $\graspregion$. The relevance of these transforms is ranked in a list $R$ according to their confidence score computed following~\eqref{eq:conf}. Therefore, the augmented library is denoted as:
\begin{equation}
    \library = \bigl\{\task_1 \rightarrow \{\pi_{\tau_1}, A_{\tau_1}, R_{\tau_1}\}, \cdots, \task_n \rightarrow \{\pi_{\tau_n}, A_{\tau_n}, R_{\tau_n}\}\bigr\}.
\end{equation}

At deployment time, we look at the spatial transform from the new grasping candidates that resembles the most well-ranked transform in $R$. This allows us to hierarchically self-assess the candidates by order of prospective success.

\subsection{Deployment on Autonomous Agent}\label{sec:framework}

Algorithm \ref{alg:framework_kb} presents the outline of \ac{SAGAT}'s end-to-end deployment, which aims at improving the success of an autonomous agent when performing a task. Given visual perception of the environment, the desired affordance, the pre-trained model to extract the grasp affordance relation (see Section~\ref{sc:kb}), the model to detect the action region, and the learnt library of task affordances (see Section~\ref{sc:mrcnn_and_dmp} to Section~\ref{sec:one_shot_library}) (lines~\ref{alg_line:input1} to \ref{alg_line:model3}), the end-to-end execution is as follows. First, the visual data is processed to extract the grasp affordance region (line~\ref{alg_line:optimal_region}) and the object's action region (line~\ref{alg_line:action_region}). The resulting grasp affordance region along with the desired affordance are used to estimate the grasp configuration proposals on the new object using the library of task affordances as prior experiences (line~\ref{alg_line:grasp_proposal}). The retrieved set of grasp configuration candidates is analysed in order of decreasing prospective success (line~\ref{alg_line:while} to line~\ref{alg_line:return_success}) until either exhausting all candidates or finding a suitable grasp for the affordance task. Importantly, the hierarchy of the proposed self-assessment analysis allows for one-shot transfer of the grasp configuration proposals, i.e. to find, on the first trial, a suitable grasp affordance by analysing the top-ranked grasp candidate. Nonetheless, the method also considers the case that exhaustive exploration of all candidates might be required, thus ensuring algorithmic completeness. 

Notably, the proposed method is not dependant on a particular grasp affordance or action region description. This modularity allows the usage of the proposed method in a wide range of setups. We demonstrate the generality of the proposed method by first, using multiple state-of-the-art approaches for grasp affordance detection, and then, determining the improvement on task performance and deployability when used altogether with our approach.

    \begin{algorithm}[t!]
        \caption{deployment of \ac{SAGAT} \label{alg:framework_kb}}
        \textbf{Input:} \\
        $\;\;$ $\text{CVF}$: camera visual feed \\ \label{alg_line:input1}
        $\;\;$ \textit{affordance}: affordance choice \\
        $\;\;$ $\texttt{graspAffordance}$: grasp affordance model \\ \label{alg_line:model1}
        $\;\;$ $\texttt{actionRegion}$: MRCNN learnt model \\ \label{alg_line:model2}
        $\;\;$ $\texttt{libTaskAffordances}$: task affordance library \\
        \label{alg_line:model3}

        \vspace{0.3em}
        \textbf{Output:} \\
        $\;\;$ $\graspoptimal$: most suitable grasp affordance configuration\\ \label{alg_line:output}
        
        \vspace{0.3em}    
        \Begin{
            \textit{$G^*$} $\gets \texttt{graspAffordance}$(CVF,$\;$\textit{affordance})\\ \label{alg_line:optimal_region}
            
            $S_O \gets$ \texttt{actionRegion}(CVF,$\;$\textit{affordance}) \\
            \label{alg_line:action_region}
            
             $\textit{$g_{p}$} \gets$ \texttt{libTaskAffordances}($G^*$,$\;$\textit{affordance})\\ \label{alg_line:grasp_proposal}
            
            \While{\KwNot \textup{\texttt{isEmpty}($g_{p}$)}} {\label{alg_line:while}
                $\grasp \gets$ \texttt{popHighestCondifence}($g_{p}$)\\ \label{alg_line:highest_confidence}
                
                $\effectnew \gets$ \texttt{forwardSimulateTask}($\grasp$, $S_O$)\\ \label{alg_line:forward_simulate}
                \If{\textup{\texttt{prospectiveTaskSuccess}($\effectnew$)}}{ \label{alg_line:prospective}
                    return $\grasp$ \label{alg_line:return_success}
                }
            }
            return none\\ \label{alg_line:next_trial}
        }
    \end{algorithm}

%% file: Sections/results.tex
\section{Experimental Evaluation and Discussion}\label{sec:results}

\begin{figure*}[th!]
  \centering\includegraphics[width= 17.7cm]{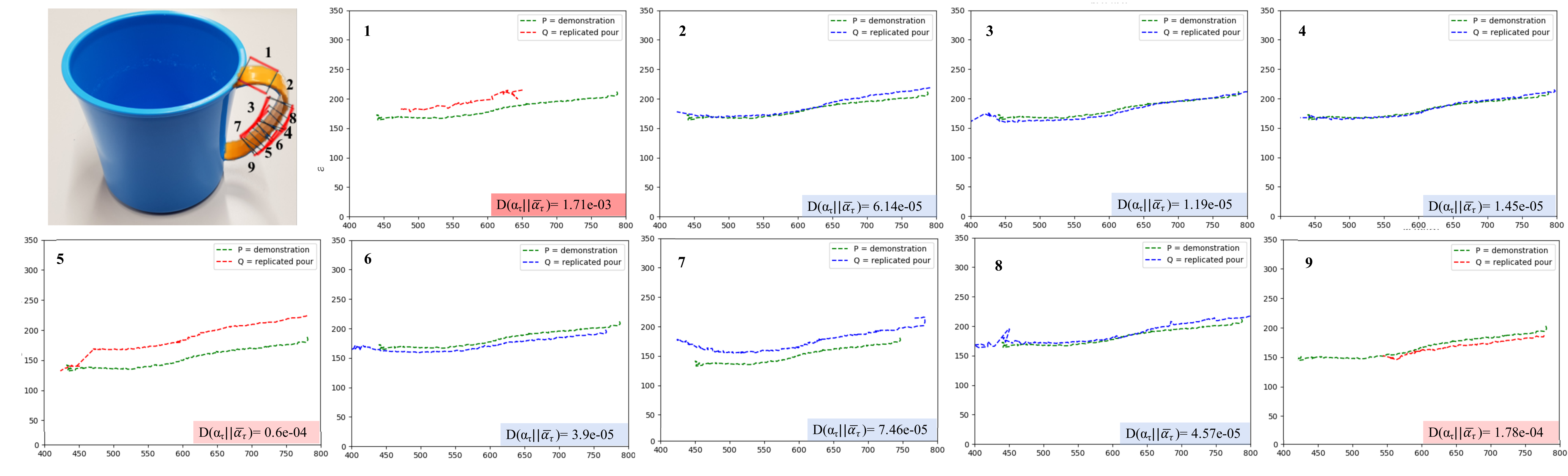}
  \caption{Entropy measurements on the \ac{2-D} frame for the pouring task. We consider as reference a socially acceptable pouring demonstration (green) against successful (blue) and undesired (red) task repetitions from different grasp candidates. The candidates are numbered with the corresponding observed effect. Successful tasks present low entropy whereas undesired effects have higher entropy. Our proposal exploits this relation to discern among grasp candidates at deployment time.
  \label{fig:examples_demo}}
\end{figure*}

The proposed methodology endows a robot with the ability to determine a suitable grasp configuration to succeed on an affordance task. Importantly, such a challenge is addressed without the need for extensive prior trials and errors. We demonstrate the potential of our method following the experimental setup described in Section~\ref{sec:experimental_setup} and a thorough evaluation based on the following tests: (i)~the spatial similarity between learnt and computed configurations across objects (Section~\ref{sc:reach_grasp}), (ii)~the accuracy of the task affordance deployment when transferred to new objects (Section~\ref{sc:zero}), and (iii)~the performance of our proposal when compared to other methodologies (Section~\ref{sc:reliability}).

\subsection{Experimental Setup \label{sec:experimental_setup}}

The end-to-end execution framework presented in Algorithm~\ref{alg:framework_kb} is deployed on a PR2 robotic platform, in both simulated and real-world scenarios. We use a Kinect mounted on the PR2's head as our visual sensor and the position sensors on the right arm joints to encode the end-effector state pose for learning the task policies in the library.

\begin{figure}[b!]
  \centering
  \includegraphics[width=6.8cm]{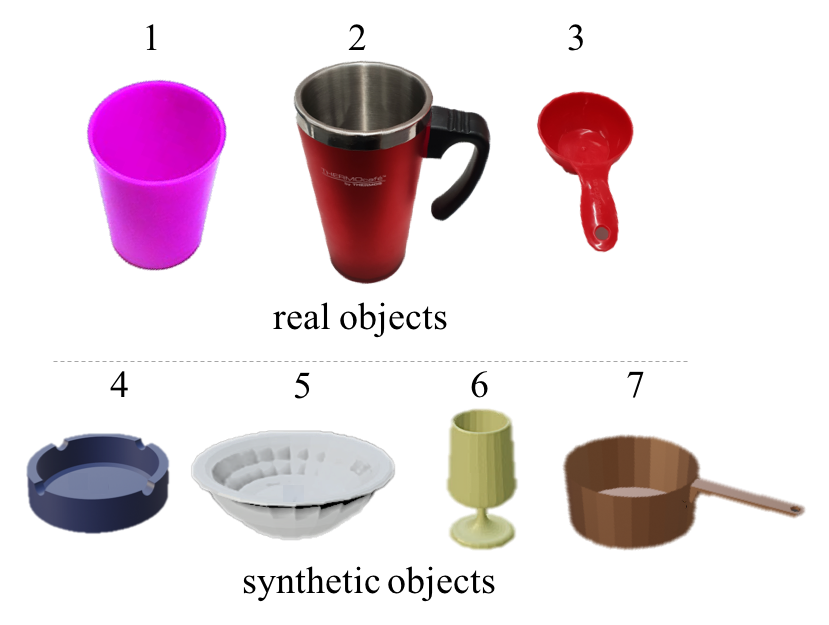}
  \caption{Novel objects to test the self-assessed grasp transfer.\label{fig:objects}}
\end{figure}

We evaluate the proposed approach with an experimental setup that considers objects with variate affordable actions and suitable grasping configurations. Particularly, the library of task affordances is built uniquely using the blue mug depicted in Fig.~\ref{fig:examples_demo}, but evaluated with the objects depicted in Fig.~\ref{fig:objects}. As can be observed, the training and testing sets present a challenging and significant variability on the grasp affordance relation. Our experimental setup also considers multiple affordances, namely: pouring, handover and shaking. The choice of these affordances is determined by those being both common among the considered objects and socially acceptable according to~\cite{ardon2019learning}.

The task policy and its expected effect corresponding to each affordance are taught to the robot via kinaesthetic demonstration. The end-effector state evolution is used to learn the task policy in form of a set of \acp{DMP}, and the state evolution of the container's action region segmented on the \ac{2-D} camera frame to learn the expected effect. As depicted in Fig.~\ref{fig:examples_demo} for the pouring task, the learnt policy is replicated $9$ times from different grasping candidates, including suitable grasp affordances (blue) and undesired deployments (red). 

The collected demonstrations are used to adjust the confidence threshold in $\eqref{eq:optimal_grasp}$ via a binary classifier, where the confidence level computed following \eqref{eq:conf} is the support, and the label $\{``successful", ``undesired"\}$ is the target. Only successful deployments are included in the library.

\begin{figure*}[ht!]
        \centering
        \subfloat[Pour ($d_{hd}=0.21$)\label{fig:pour_d_h}]{
            \centering    \includegraphics[width=5.85cm]{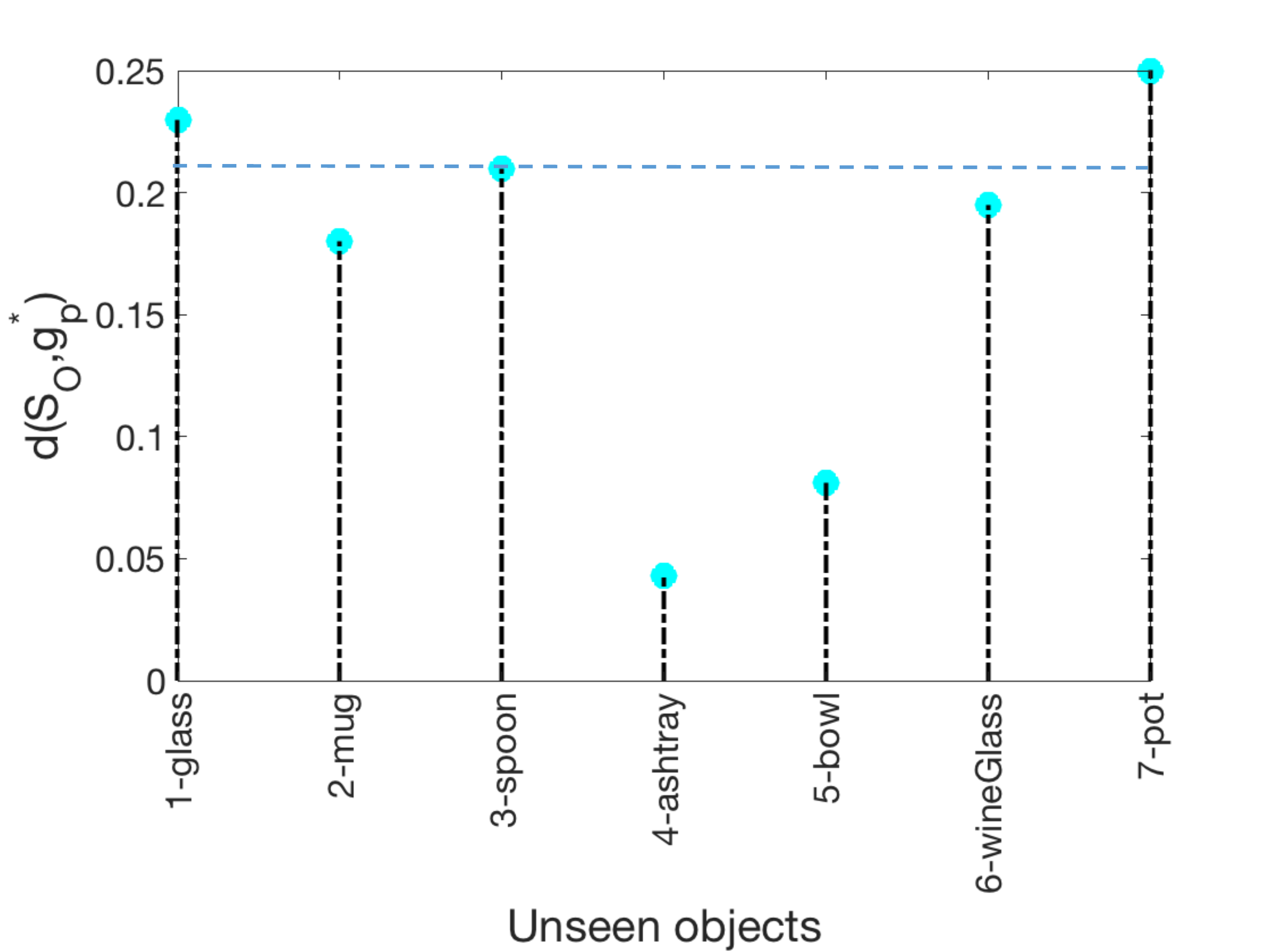}
        } \hspace{-0.4cm}
        \subfloat[Shake ($d_{hd}=0.23$)\label{fig:shake_d_h} ]{
            \centering                \includegraphics[width=5.85cm]{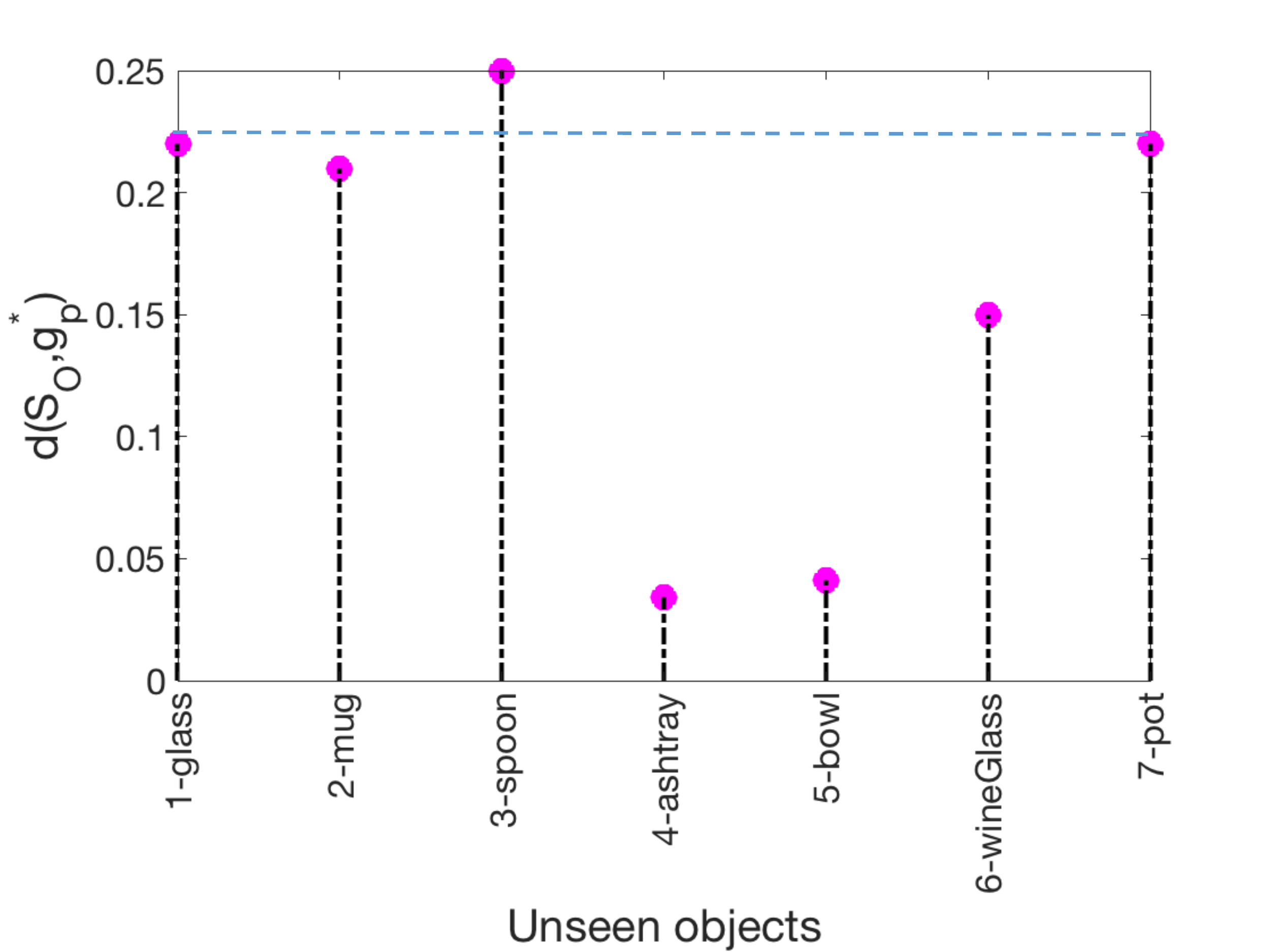}
        } \hspace{-0.45cm}
        \subfloat[Handover ($d_{hd}=0.20$)\label{fig:handover_d_h}]{
            \centering
        \includegraphics[width=5.85cm]{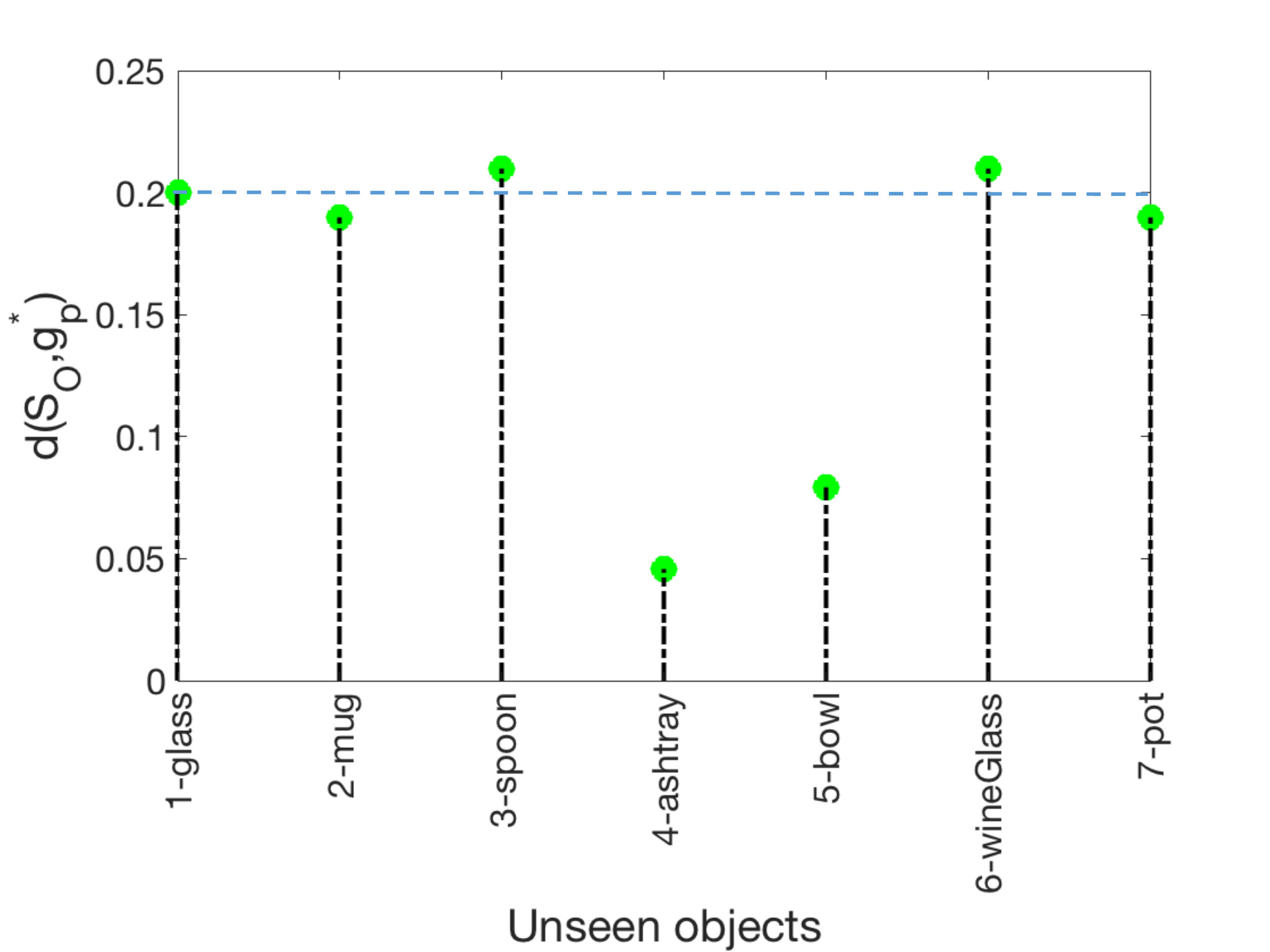}
        }
       \caption{Visualisation of the dissimilarity metric between an object's action region and the corresponding suitable grasp configuration, in comparison to the mean dissimilarity observed during the demonstrations ($d_{hd}$, blue horizontal line).}
            \label{fig:object_samples_for_zero_shot}
    \end{figure*}


\subsection{Spatial Similarity of Grasp Configurations\label{sc:reach_grasp}}

Our method allows the system for one-shot transfer of grasp configurations to new objects. As explained in Section~\ref{sec:one_shot_library}, we rank the grasp candidates on new objects as those that closely resemble the experiences stored in the library of task affordances. This approximation is based on the expectation that similar spatial configurations should offer similar performance when dealing with the same task. In this set of experiments, we demonstrate the validity of such a hypothesis by evaluating the spatial similarity between the proposals estimated on new objects and the ones previously identified as suitable and stored in the library.

For an object, we calculate the Euclidean distance
between the segmented action region $S_O$ and the obtained grasp configuration $\graspoptimal$. Fig.~\ref{fig:object_samples_for_zero_shot} shows the obtained distances denoted as $d_h(S_O,\graspoptimal)$. The blue horizontal line represents the mean distance obtained during the demonstrations.
Overall, we observe similar distances from action regions to grasp configurations across objects. For dissimilar cases such as $4$ and $5$ (ashtray and bowl respectively), the difference is given by the fact that the obtained grasping region for most of the tasks lies on the edges of the object compartment. Even though these grasping configurations are relatively close to the action region, we will see on Table~\ref{tb:grasp} that the average performance of the tasks is preserved.

\begin{figure}[b!]
  \centering
  \includegraphics[width=7cm]{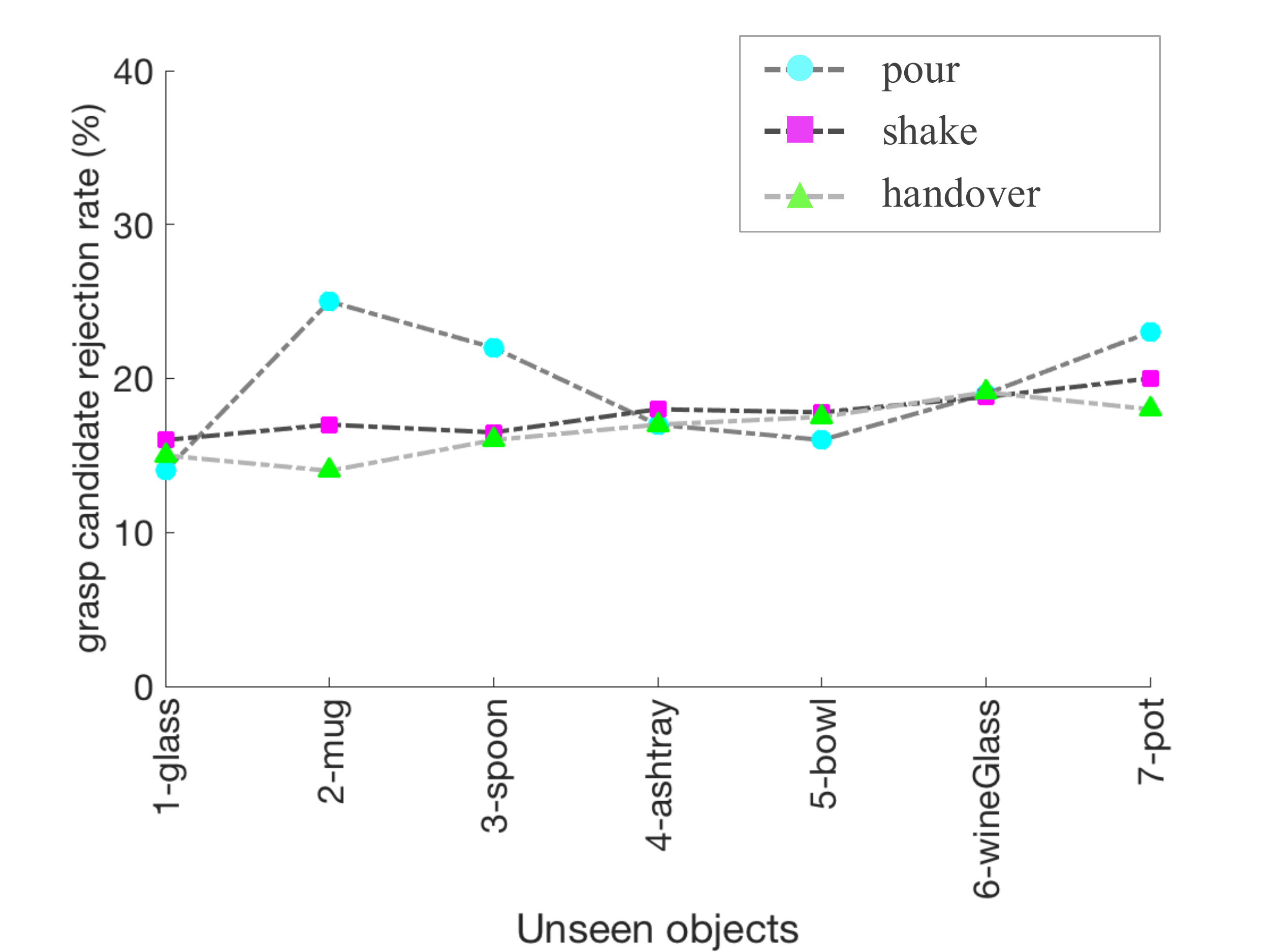}
  \caption{
  Rejection rate of grasp candidates with prospective unsuccessful task deployment. Grasp configurations, as extracted with DeepGrasp \cite{chu2018real}, that do not relate to the prior on successful task deployment, as stored in the library, are rejected in the one-shot transfer scheme\label{fig:rejection}.
  }
\end{figure}

To further evaluate similarity across obtained grasping configurations, we are also interested in how much the system prunes the grasping space based on the information stored in the library. 
As defined in~\eqref{eq:optimal_grasp}, we use a confidence threshold for the pruning process of the grasping space. Thus, based on the prior of well-performing grasp configurations, highly dissimilar proposals are not considered on the self-assessed transfer process.
Fig.~\ref{fig:rejection} depicts the rejection rate of grasp configuration proposals per task affordance. From the plot, we see that the pouring task shows the highest rejection rate, especially for objects that have handles. This hints that for this task the grasping choice is more critical.

\subsection{One-Shot Transfer of Task Affordances\label{sc:zero}}

\begin{figure}[bh!]
        \centering
        \subfloat[Pour task affordance\label{fig:pour}]{
            \centering    \includegraphics[width=8.4cm]{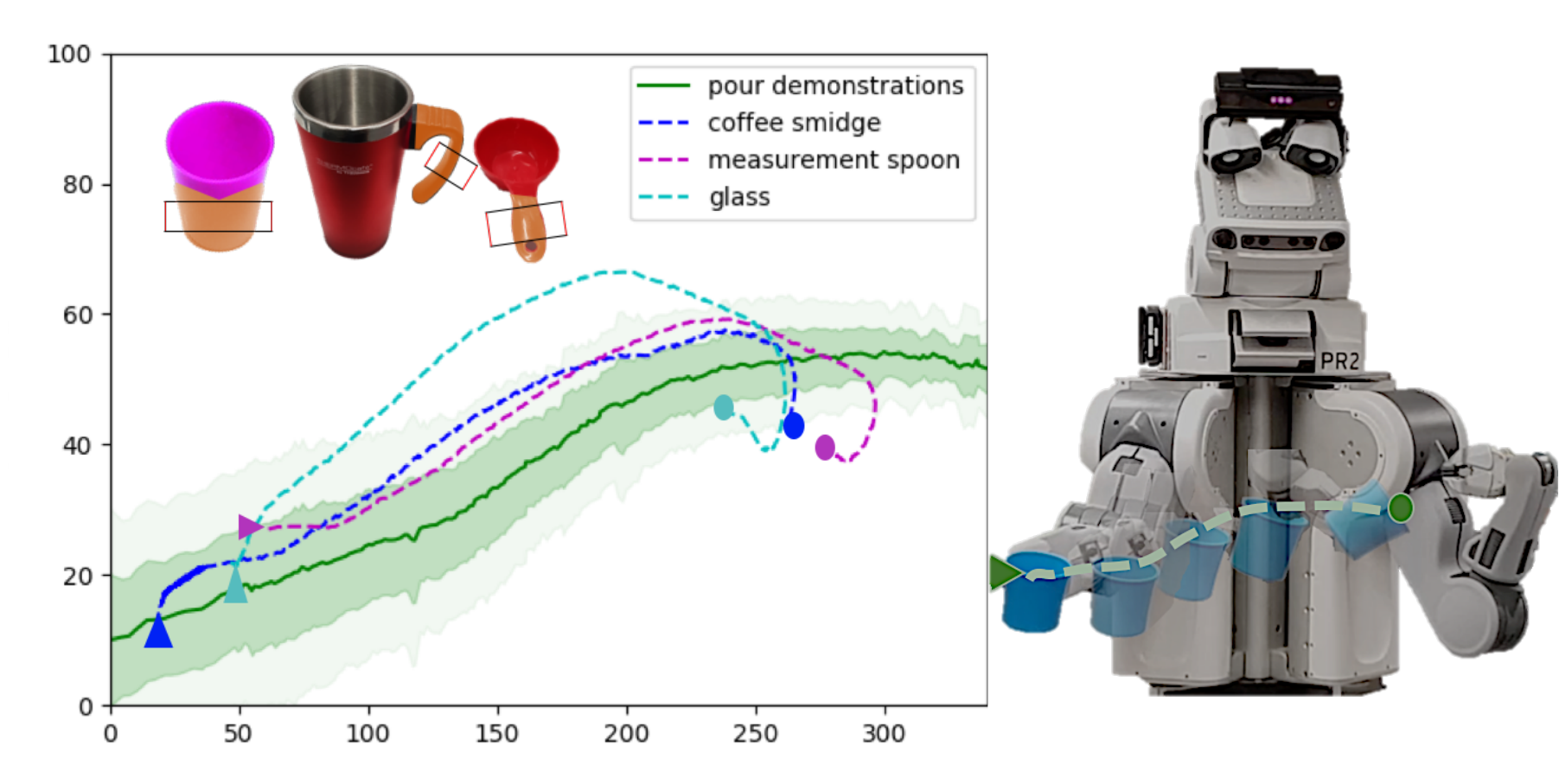}
        } 
        
        \subfloat[Shake task affordance\label{fig:shake} ]{
            \centering
                \includegraphics[width=8.4cm]{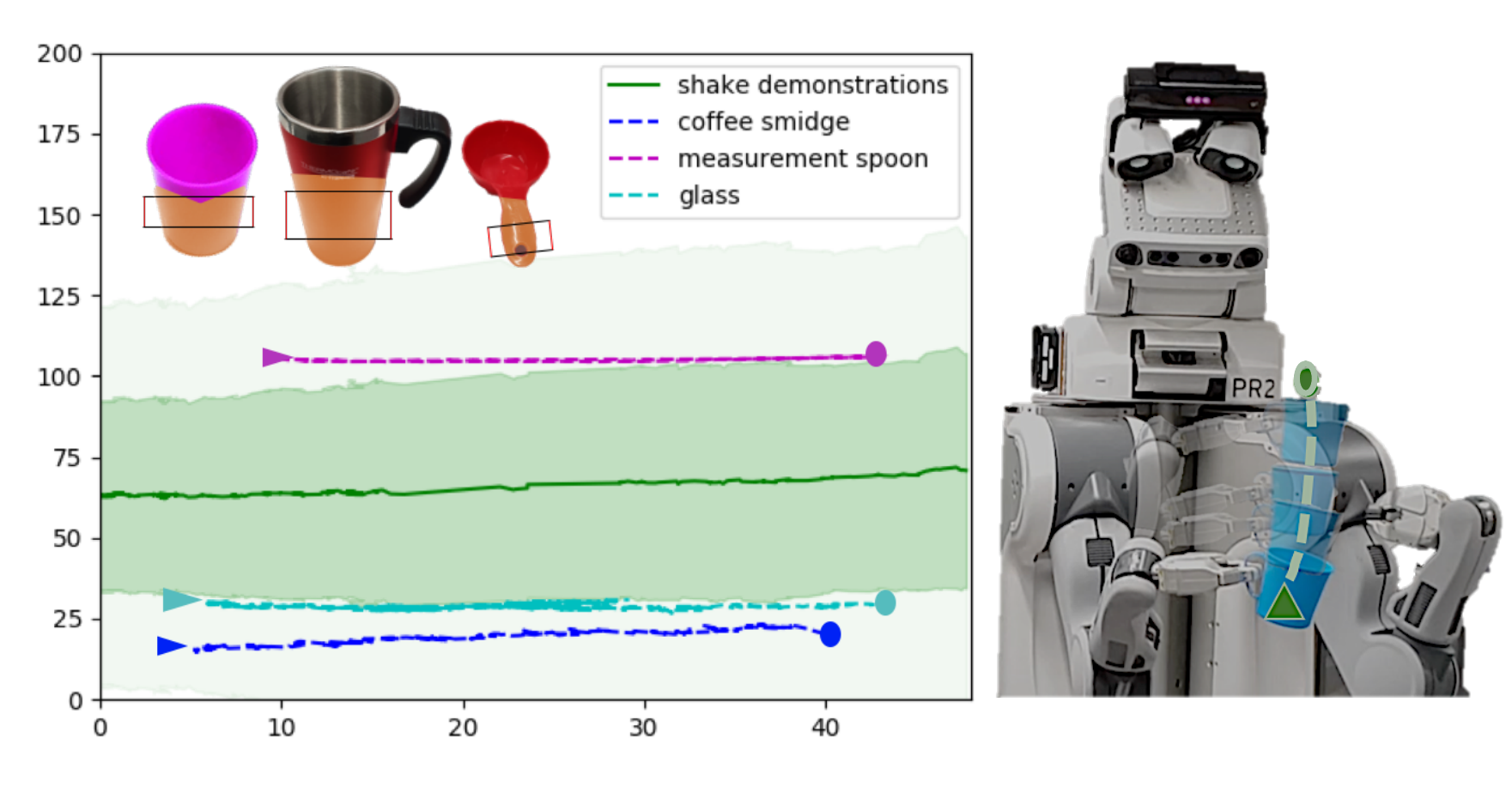}
        }
        
        \subfloat[Handover task affordance\label{fig:handover}]{
            \centering
        \includegraphics[width=8.4cm]{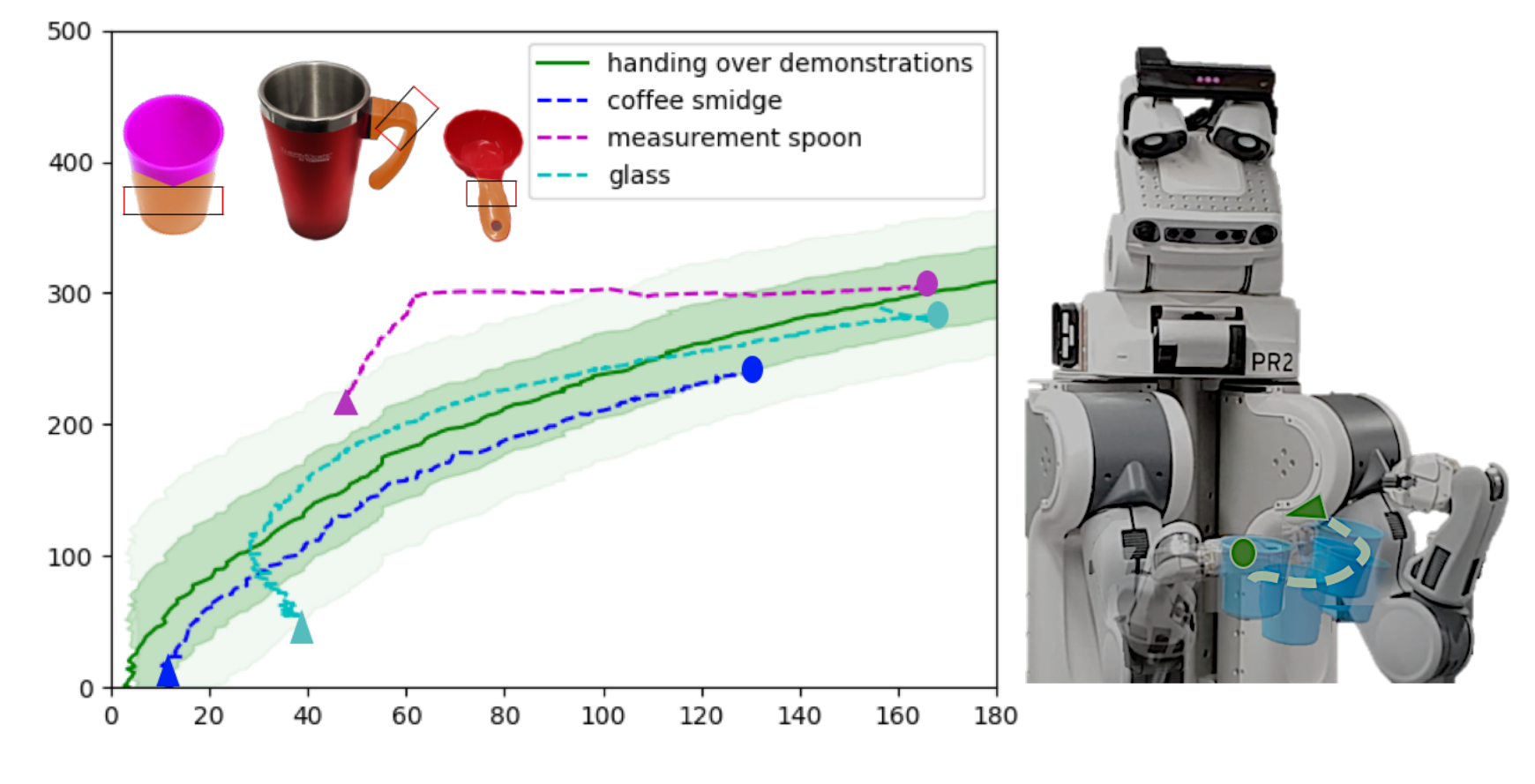}
        }
       \caption{Task affordance performance when deployed on novel objects (colour-coded lines) in comparison with the multiple successful demonstrations (green scale distribution).}
            \label{fig:trajectories}
    \end{figure}

The second experimental test analyses the performance of our method when addressing task affordances on new objects. The goal of this evaluation is to determine if the chosen grasp configuration enables objects to perform the task affordance as successfully as the prior stored in the library. Fig.~\ref{fig:trajectories} depicts the mean and variance (green scale) of the prior experiences in the library for the tasks pour, shake and handover. Each task was performed with three real objects with notably different features: a travel mug (dark blue), measurement spoon (magenta) and a glass (blue). The resulting effect when performing the tasks from the computed grasping configuration is colour-coded on top of the prior experiences distribution.

Subject to the task affordance, the three objects show different grasp affordance regions. After the one-shot self-assessment procedure, the computed grasp configurations are the most spatially similar to the most successful grasp configuration in the experience dataset. Importantly, as illustrated in Fig.~\ref{fig:trajectories}, this strategy is invariant to different initial and final states of the task. This is reflected in the obtained task affordance effect, which falls inside the variance of the demonstrations.


\subsection{Comparison of Task Deployment Reliability\label{sc:reliability}}
     
    \begin{figure}[t!]
      \centering
      \includegraphics[width=8.7cm]{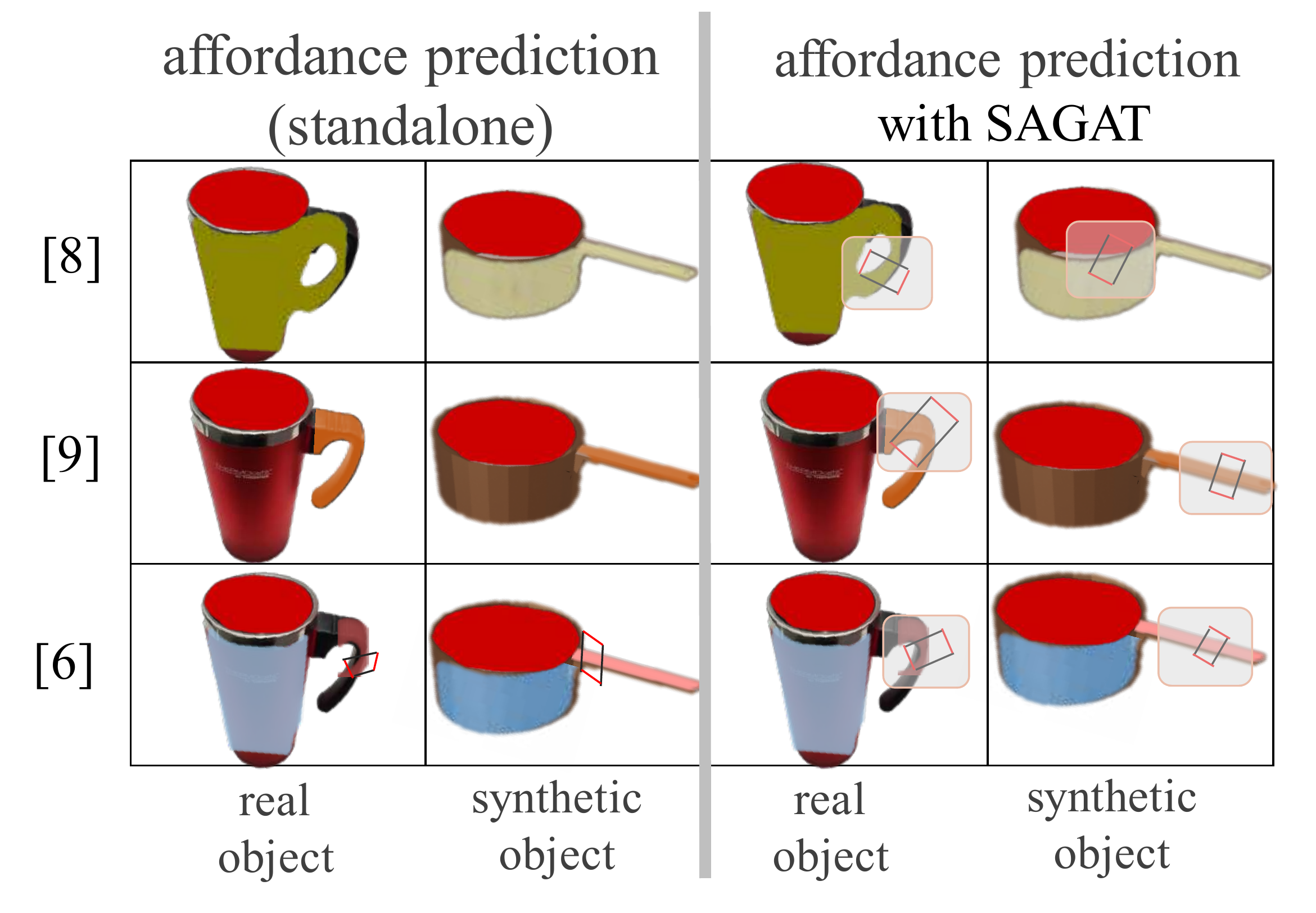}
      \caption{Comparison of grasp affordance detection for the task of pouring with state-of-the-art methods and \ac{SAGAT}. The resulting grasp configuration proposals obtained with \ac{SAGAT} are highlighted for better visualisation. \label{fig:soa_comparison}}
    \end{figure}

The last experimental test is to demonstrate at which level the proposed method enhances the task deployment reliability when used in conjunction with methods for grasp affordance detection~\cite{AffordanceNet18,chu2019learning,ardon2019learning}. To conduct this evaluation, we use the open-source implementations of~\cite{AffordanceNet18,chu2019learning,ardon2019learning} on all objects illustrated in Fig.~\ref{fig:objects}, in the real and simulated robotic platform. The obtained grasp regions are used to execute the task in two different ways: (i)~in stand-alone fashion, i.e. as originally proposed, and (ii)~as input of our \ac{SAGAT} approach to determine the most suitable grasp candidate. Fig.~\ref{fig:soa_comparison} shows some examples of the grasp affordance detected with the previously mentioned methods and our approach.

We use the policies in the learnt library of task affordances to replicate the pour, shake and handover tasks on each object, for each grasp affordance, and for each method when used as stand-alone and combined with \ac{SAGAT}. This results in a total of $126$ tasks deployments on the robotic platform\footnote{A compilation of experiments can be found in: \url{https://youtu.be/nCCc3_Rk8Ks}}. Table~\ref{tb:grasp} summarises the obtained results. As can be observed, deploying a task using state-of-the-art methods on grasp affordance detection provides an average success rate of $79.2\%$ across tasks. With our approach, the deployability success is enhanced for all the tasks, with an average rate of $85.4\%$. Interestingly, the $5.2\%$ improvement is not equally distributed across tasks; more challenging tasks experience a higher success rate. This is the case of the pouring tasks where deployability success is increased by $11.67\%$.

    \begin{table}[t!]
        \renewcommand{\arraystretch}{1.25}
         \centering
         \begin{tabular}{|p{0.8cm}|c|c|c|c|c|c|}
      \cline{2-7}
      
      \multicolumn{1}{c|}{} &  \textbf{\notsotiny{\cite{AffordanceNet18}}} & \!\!\textbf{\notsotiny{\cite{AffordanceNet18}$+$}\ours\!\!} & \textbf{\notsotiny{\cite{ardon2019learning}}} & \!\!\textbf{\notsotiny{\cite{ardon2019learning}$+$}\ours\!\!} & \textbf{\notsotiny{\cite{chu2019learning}}} & \!\!\textbf{\notsotiny{\cite{chu2019learning}$+$}\ours\!\!} \\ 
      
      \hline
     
      \!\!\!Pour & 70\% & \textbf{82\%} &72\% & \textbf{83\% } & 73\% & \textbf{85\%}\\ \hline 
      \!\!\!Shake & 84\%  & \textbf{87\%} & 85\% & \textbf{87\%} & 86\% & \textbf{88\% }\\ \hline
      \!\!\!Handover & 80\% & \textbf{85\%} & 81\% &   \textbf{86\%} & 82\% & \textbf{86\%} \\ \hline
      
         \end{tabular}
         \caption{Comparison of success rates on task affordance deployment when using state-of-the-art grasp affordance extractors as stand-alone and with our method.\label{tb:grasp}}
     \end{table}    

%% file: Sections/contribution.tex
\section{Conclusions and Future Work}\label{sec:contribution}


In this paper, we presented a novel experience-based pipeline for \acf{SAGAT}. Our approach enhances the deployment reliability of current state-of-the-art methods on grasp affordance detection, by extracting multiple grasp configuration candidates from a given grasp affordance region. The outcome of executing a task from different grasp candidates is estimated via forward simulation. These estimates are evaluated and ranked via a heuristic confidence function in relation to task performance and grasp configuration candidates. Such information is stored in a library of task affordances, which serves as a basis for one-shot transfer estimation to identify grasp affordance configurations similar to those previously experienced, with the insight that similar regions lead to similar deployments of the task. We evaluate the method's efficacy on novel task affordance problems by training on a single object and testing on multiple new ones. We observe a significant performance improvement up to approximately $11.7\%$ in our experiments when using our proposal in comparison to state-of-the-art approaches on grasp affordance detection. Experimental evaluation on a PR2 robotic platform demonstrates highly reliable deployability of the proposed method to deal with real-world task affordance problems. 

This work encourages multiple interesting directions for future work. Our follow-up work will study a unified probabilistic framework to infer the most suitable grasp affordance candidate. We envision that this will allow sets of actions and grasps to be predicted when dealing with multiple correlated objects in the scene. Another interesting extension is the assessment of the end-state comfort-effect for grasping in human-robot collaboration tasks, such that the robot's grasp affordance considers the human's grasp capabilities.

%% file: main.bbl
\begin{thebibliography}{10}

\bibitem{jamone2018affordances}
L.~Jamone, E.~Ugur, A.~Cangelosi, L.~Fadiga, A.~Bernardino, J.~Piater, and
  J.~Santos-Victor, ``Affordances in psychology, neuroscience, and robotics: A
  survey,'' {\em IEEE Transactions on Cognitive and Developmental Systems},
  vol.~10, no.~1, pp.~4--25, 2018.

\bibitem{min2016affordance}
H.~Min, C.~Yi, R.~Luo, J.~Zhu, and S.~Bi, ``Affordance research in
  developmental robotics: a survey,'' {\em IEEE Transactions on Cognitive and
  Developmental Systems}, vol.~8, no.~4, pp.~237--255, 2016.

\bibitem{Gibson77-affordances}
J.~Gibson, ``{T}he theory of affordances,'' in {\em {P}erceiving, {A}cting, and
  {K}nowing: {T}oward and {E}cological {P}sychology} (R.~Shaw and J.~Bransford,
  eds.), pp.~62--82, Hillsdale, NJ: {E}rlbaum, 1977.

\bibitem{Lenz2015}
I.~Lenz, H.~Lee, and A.~Saxena, ``{Deep learning for detecting robotic
  grasps},'' {\em International Journal of Robotics Research}, vol.~34,
  no.~4-5, pp.~705--724, 2015.

\bibitem{chu2018real}
F.-J. Chu, R.~Xu, and P.~A. Vela, ``Real-world multiobject, multigrasp
  detection,'' {\em IEEE Robotics and Automation Letters}, vol.~3, no.~4,
  pp.~3355--3362, 2018.

\bibitem{chu2019learning}
F.-J. Chu, R.~Xu, and P.~A. Vela, ``Learning affordance segmentation for
  real-world robotic manipulation via synthetic images,'' {\em IEEE Robotics
  and Automation Letters}, vol.~4, no.~2, pp.~1140--1147, 2019.

\bibitem{bohg2010learning}
J.~Bohg and D.~Kragic, ``Learning grasping points with shape context,'' {\em
  Robotics and Autonomous Systems}, vol.~58, no.~4, pp.~362--377, 2010.

\bibitem{AffordanceNet18}
T.-T. Do, A.~Nguyen, and I.~Reid, ``Affordancenet: An end-to-end deep learning
  approach for object affordance detection,'' in {\em International Conference
  on Robotics and Automation (ICRA)}, 2018.

\bibitem{ardon2019learning}
P.~Ard{\'o}n, {\`E}.~Pairet, R.~P. Petrick, S.~Ramamoorthy, and K.~S. Lohan,
  ``Learning grasp affordance reasoning through semantic relations,'' {\em IEEE
  Robotics and Automation Letters}, vol.~4, no.~4, pp.~4571--4578, 2019.

\bibitem{Montesano2008LearningOA}
L.~Montesano, M.~Lopes, A.~Bernardino, and J.~Santos-Victor, ``Learning object
  affordances: From sensory--motor coordination to imitation,'' {\em IEEE
  Trans. Robotics}, vol.~24, pp.~15--26, 2008.

\bibitem{fang2019learning}
K.~Fang, Y.~Zhu, A.~Garg, A.~Kurenkov, V.~Mehta, L.~Fei-Fei, and S.~Savarese,
  ``Learning task-oriented grasping for tool manipulation from simulated
  self-supervision,'' {\em The International Journal of Robotics Research},
  p.~0278364919872545, 2019.

\bibitem{mandlekar2018roboturk}
A.~Mandlekar, Y.~Zhu, A.~Garg, J.~Booher, M.~Spero, A.~Tung, J.~Gao, J.~Emmons,
  A.~Gupta, E.~Orbay, {\em et~al.}, ``Roboturk: A crowdsourcing platform for
  robotic skill learning through imitation,'' in {\em Conference on Robot
  Learning}, pp.~879--893, 2018.

\bibitem{kroemer2012kernel}
O.~Kroemer, E.~Ugur, E.~Oztop, and J.~Peters, ``A kernel-based approach to
  direct action perception,'' in {\em 2012 IEEE International Conference on
  Robotics and Automation}, pp.~2605--2610, IEEE, 2012.

\bibitem{kruger2011object}
N.~Kr{\"u}ger, C.~Geib, J.~Piater, R.~Petrick, M.~Steedman,
  F.~W{\"o}rg{\"o}tter, A.~Ude, T.~Asfour, D.~Kraft, D.~Omr{\v{c}}en, {\em
  et~al.}, ``Object--action complexes: Grounded abstractions of sensory--motor
  processes,'' {\em Robotics and Autonomous Systems}, vol.~59, no.~10,
  pp.~740--757, 2011.

\bibitem{song2010learning}
D.~Song, K.~Huebner, V.~Kyrki, and D.~Kragic, ``Learning task constraints for
  robot grasping using graphical models,'' in {\em Intelligent Robots and
  Systems (IROS), 2010 IEEE/RSJ International Conference on}, pp.~1579--1585,
  IEEE, 2010.

\bibitem{Montesano2009LearningGA}
L.~Montesano and M.~Lopes, ``Learning grasping affordances from local visual
  descriptors,'' in {\em Development and Learning, 2009. ICDL 2009. IEEE 8th
  International Conference on}, pp.~1--6, IEEE, 2009.

\bibitem{moldovan2012learning}
B.~Moldovan, P.~Moreno, M.~van Otterlo, J.~Santos-Victor, and L.~De~Raedt,
  ``Learning relational affordance models for robots in multi-object
  manipulation tasks,'' in {\em Robotics and Automation (ICRA), 2012 IEEE
  International Conference on}, pp.~4373--4378, IEEE, 2012.

\bibitem{pairet2019learning}
{\`E}.~Pairet, P.~Ard{\'o}n, M.~Mistry, and Y.~Petillot, ``Learning and
  composing primitive skills for dual-arm manipulation,'' in {\em Annual
  Conference Towards Autonomous Robotic Systems}, pp.~65--77, Springer, 2019.

\bibitem{ijspeert2013dynamical}
A.~J. Ijspeert, J.~Nakanishi, H.~Hoffmann, P.~Pastor, and S.~Schaal,
  ``Dynamical movement primitives: learning attractor models for motor
  behaviors,'' {\em Neural computation}, vol.~25, no.~2, pp.~328--373, 2013.

\bibitem{pairet2019learningb}
{\`E}.~Pairet, P.~Ard{\'o}n, M.~Mistry, and Y.~Petillot, ``Learning
  generalizable coupling terms for obstacle avoidance via low-dimensional
  geometric descriptors,'' {\em IEEE Robotics and Automation Letters}, vol.~4,
  no.~4, pp.~3979--3986, 2019.

\bibitem{he2017mask}
K.~He, G.~Gkioxari, P.~Doll{\'a}r, and R.~Girshick, ``Mask r-cnn,'' in {\em
  Proceedings of the IEEE international conference on computer vision},
  pp.~2961--2969, 2017.

\bibitem{perez2008kullback}
F.~P{\'e}rez-Cruz, ``Kullback-leibler divergence estimation of continuous
  distributions,'' in {\em 2008 IEEE international symposium on information
  theory}, pp.~1666--1670, IEEE, 2008.

\end{thebibliography}
